\documentclass[a4paper,fleqn]{cas-sc}

\usepackage[authoryear,longnamesfirst]{natbib}
\usepackage{etoolbox}
\usepackage{xspace}
\usepackage{siunitx}

\def\tsc#1{\csdef{#1}{\textsc{\lowercase{#1}}\xspace}}
\tsc{WGM}
\tsc{QE}
\tsc{EP}
\tsc{PMS}
\tsc{BEC}
\tsc{DE}

\begin{document}
\let\WriteBookmarks\relax
\def\floatpagepagefraction{1}
\def\textpagefraction{.001}

\shorttitle{AlignVTOFF: Texture-Spatial Feature Alignment for High-Fidelity Virtual Try-Off}
\shortauthors{Yihan Zhu and Mengying Ge}

\title [mode = title]{AlignVTOFF: Texture-Spatial Feature Alignment for High-Fidelity Virtual Try-Off}

\author[1]{Yihan Zhu}
\author[2]{Mengying Ge\corref{cor1}}[orcid=0009-0007-0361-3194]
\ead{gemengying@shu.edu.cn}
\cormark[1] 

\address[1]{Sino-European School of Technology, Shanghai University, 99 Shangda Road, Baoshan District, Shanghai 200444, China}

\address[2]{National Demonstration Center for Experimental Engineering Training Education, Shanghai University, 99 Shangda Road, Baoshan District, Shanghai 200444, China}

\cortext[cor1]{Corresponding author at: National Demonstration Center for Experimental Engineering Training Education, Shanghai University, 99 Shangda Road, Baoshan District, Shanghai 200444, China.}

\begin{abstract}
Virtual Try-Off (VTOFF) is a challenging multimodal image generation task that aims to synthesize high-fidelity flat-lay garments under complex geometric deformation and rich high-frequency textures. Existing methods often rely on lightweight modules for fast feature extraction, which struggles to preserve structured patterns and fine-grained details, leading to texture attenuation during generation.
To address these issues, we propose AlignVTOFF, a novel parallel U-Net framework built upon a Reference U-Net and Texture–Spatial Feature Alignment (TSFA). The Reference U-Net performs multi-scale feature extraction and enhances geometric fidelity, enabling robust modeling of deformation while retaining complex structured patterns. TSFA then injects the reference garment features into a frozen denoising U-Net via a hybrid attention design, consisting of a trainable cross-attention module and a frozen self-attention module. This design explicitly aligns texture and spatial cues and alleviates the loss of high-frequency information during the denoising process.
Extensive experiments across multiple settings demonstrate that AlignVTOFF consistently outperforms state-of-the-art methods, producing flat-lay garment results with improved structural realism and high-frequency detail fidelity.

\end{abstract}


\maketitle

\section{Introduction}
This paper studies Virtual Try-Off (VTOFF), the task of extracting canonical flat-lay garment images from photos of clothed individuals~\cite{velioglu2025tryoffdiffvirtualtryoffhighfidelitygarment}. VTOFF is highly relevant to practical e-commerce applications, as it complements Virtual Try-On (VTON)~\cite{10.1145/3447239,ding2023computational,gao2024exploring} and Virtual Dressing (VD)~\cite{shen2025imagdressing} by enabling standardized garment representation for downstream retrieval, editing, and interactive shopping experiences. 

Early VTON systems were predominantly based on generative adversarial networks (GANs)~\cite{8253599}, typically consisting of a warping module that learns correspondences between garments and human bodies, and a generator module that synthesizes the target garment on the person. Despite progress, GAN-based approaches~\cite{han2018viton,choi2021viton} often suffer from training instability~\cite{arjovsky2017wasserstein}, limited ability to preserve fine-grained textures, and reduced robustness to complex backgrounds and occlusions~\cite{zhu2023tryondiffusion}. 

With the rise of diffusion models~\cite{ramesh2022hierarchical,rombach2022high}, VTON and VD have witnessed rapid advances, benefiting from strong generative priors and iterative denoising that better supports texture synthesis and high perceptual fidelity. Recent diffusion-based pipelines further extend controllability by incorporating richer conditions, such as pose, layout, and semantic constraints, to improve structural consistency and user control~\cite{shen2024advancing,shen2024imagpose,shen2025boosting,shen2025imagharmony,shen2025imagedit}. Meanwhile, fine-grained garment generation has also progressed toward controllable fashion design and high-frequency detail modeling~\cite{shen2025imaggarment}. 
However, a major bottleneck for both VTOFF and downstream VD and VTON pipelines is the scarcity of large-scale, high-quality canonical garment data paired with in-the-wild person images. This motivates the development of high-fidelity generative models that can reconstruct clearer and more standardized flat-lay garments to facilitate effective feature reuse and interaction in multimodal pipelines.

\begin{figure}
	\centering
	\includegraphics[width=0.9\linewidth]{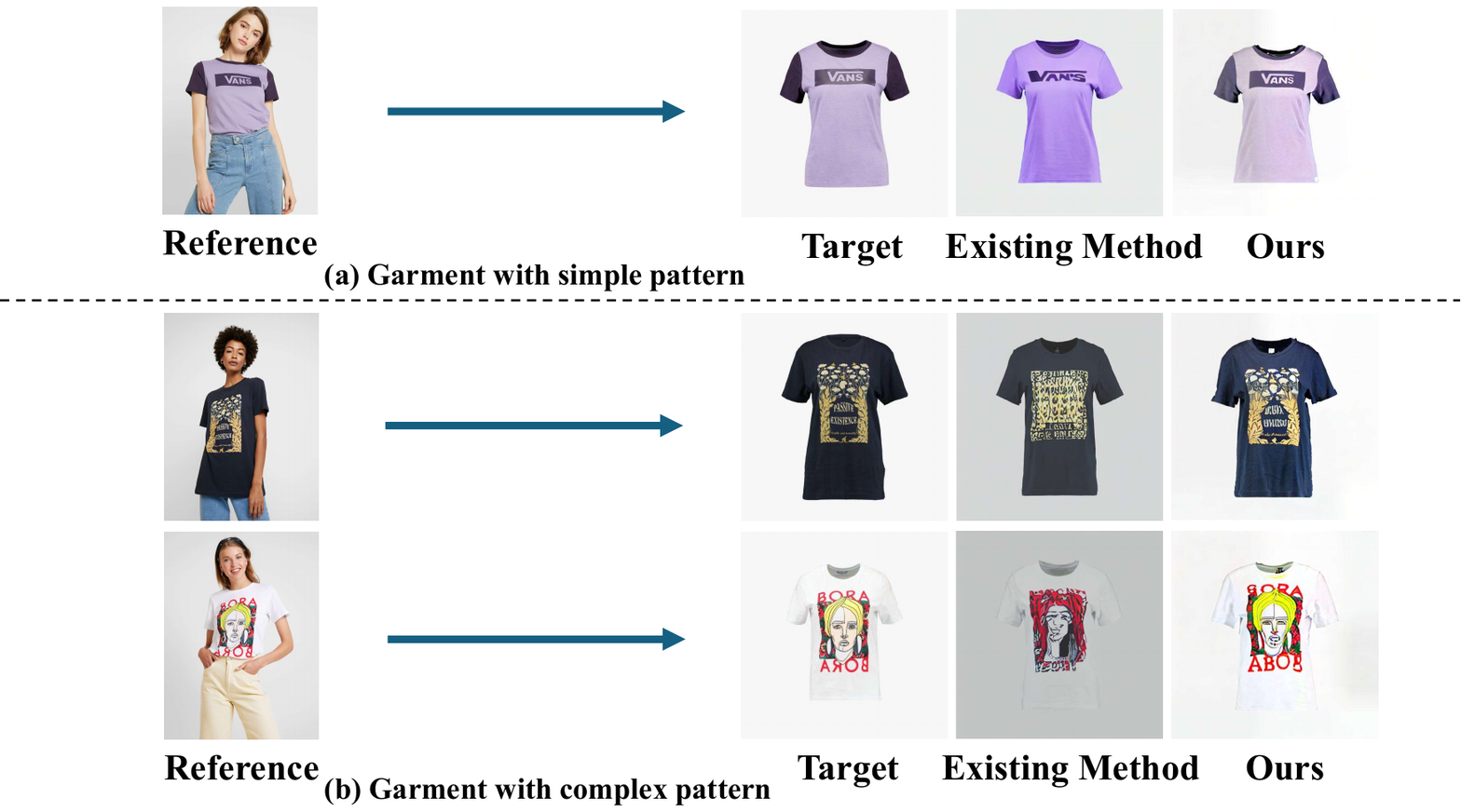}
	\caption{VTOFF results on (a) common garment and (b) garment with complex pattern. Existing methods struggle to preserve the
complex structured patterns, while ours ensures consistent and faithful generation.}
	\label{fig:1}
\end{figure}

Considering VTOFF as an inverse generation problem to VTON, recent works have adopted the latent diffusion model (LDM) paradigm. In particular, TryOffDiff~\cite{velioglu2025tryoffdiffvirtualtryoffhighfidelitygarment} encodes the clothed person image using SigLIP~\cite{zhai2023sigmoid} and CLIP, and injects the extracted representations into the denoising U-Net via a customized Adapter module~\cite{ye2023ip}. 
However, we argue that such a single-stream architecture introduces an inherent mismatch with the requirements of VTOFF: the standard downsampling path behaves like a low-pass filter, progressively discarding the high-frequency cues that are essential for faithful garment reconstruction. This issue is especially severe for garments with complex structured patterns, where existing methods tend to produce blurred textures and distorted geometry (see Fig.~\ref{fig:1}(b)).

To address this limitation, we propose a novel parallel U-Net framework, termed Texture-Spatial Alignment for Virtual Try-Off (AlignVTOFF). AlignVTOFF is a latent diffusion model designed to bypass the information bottleneck caused by standard downsampling. Leveraging the near-lossless reconstruction capability of VAE latents, we introduce a dedicated Reference U-Net that jointly captures semantic cues from CLIP and texture cues from the VAE, thereby preserving high-frequency details that are otherwise attenuated in conventional encoding paths. The extracted reference features are then integrated into a frozen Denoising U-Net via Texture-Spatial Feature Alignment (TSFA), which combines a frozen self-attention branch with a trainable cross-attention branch to align texture and spatial structures during denoising. Moreover, AlignVTOFF is compatible with external conditional extensions such as ControlNet~\cite{zhang2023adding}, enabling enhanced generation diversity and controllability without redesigning the core architecture.

The main contributions of this paper are summarized as follows:
\begin{itemize}
    \item We formulate VTOFF as a latent diffusion problem tailored for generating standardized canonical garment images from human-worn garment inputs.
    \item We propose AlignVTOFF, featuring a dedicated Reference U-Net to extract complementary semantic cues from CLIP and texture cues from the VAE, and a TSFA module to inject implicit geometric clues from the reference image while preserving essential garment appearance details within a frozen denoising backbone.
    \item We demonstrate that AlignVTOFF can be seamlessly combined with external conditional extensions, such as ControlNet, to further improve controllability and output diversity.
    \item Extensive experiments on VITON-HD and Dress Code show that AlignVTOFF achieves state-of-the-art performance, substantially improving structural accuracy and high-frequency texture preservation.
\end{itemize}

\section{Related Work}\label{sec:rw}

\subsection{Virtual Try-On and Virtual Try-Off}
As a counterpart to Virtual Try-Off (VTOFF), Virtual Try-On (VTON) aims to synthesize an image in which a target garment is naturally transferred to the corresponding region of a person~\cite{han2018viton}, while preserving identity, pose, and body shape and maintaining fine garment details. Early VTON systems were predominantly built on Generative Adversarial Networks (GANs)~\cite{dong2024internlm,jo2019sc,lee2020maskgan}, such as FW-GAN~\cite{dong2019fw} and PASTA-GAN~\cite{xie2021towards}. A representative example is VITON~\cite{han2018viton}, which follows a two-stage pipeline: it first warps the target garment to match the desired body shape and then applies a GAN-based generator to blend the warped garment onto the person. Despite their effectiveness, GAN-based methods are often hindered by training instability induced by adversarial min--max optimization, and they tend to struggle with complex backgrounds and the preservation of high-frequency textures.

Recent research has increasingly shifted toward diffusion-based priors for VTON, yielding improved stability and better detail retention. Methods such as LADI-VTON~\cite{morelli2023ladi}, DC-VTON~\cite{gou2023taming}, CatVTON~\cite{chong2024catvton}, and OOTDiffusion~\cite{xu2025ootdiffusion} demonstrate that diffusion models can generate higher-quality try-on results with fewer artifacts. In contrast, VTOFF poses an inverse challenge: the model must reconstruct a standardized, canonical garment from a real photograph of a clothed person. Early attempts such as TileGAN~\cite{zeng2020tilegan} adopt a two-stage GAN framework but do not formalize VTOFF as a standalone research problem and inherit the limitations of GAN training. Subsequent diffusion-based garment synthesis approaches, including SGDiff~\cite{sun2023sgdiff} and DiffCloth~\cite{zhang2023diffcloth}, leverage pretrained diffusion backbones (e.g., GLIDE or Stable Diffusion) and typically rely on multimodal inputs such as text with style images or sketches; however, these settings differ from VTOFF, where the input is a real-world clothed-person photo and reconstruction fidelity is central. VTOFF also relates to image attribute manipulation and fashion editing, where methods such as AIRR~\cite{li2022supervised} and VPTNet~\cite{kwon2022tailor} disentangle attributes (e.g., color, shape, appearance) to enable controllable edits, but they often assume aligned or clean images and thus under-address the deformations and occlusions found in real photos. As a pioneer, TryOffDiff~\cite{velioglu2025tryoffdiffvirtualtryoffhighfidelitygarment} introduces VTOFF by adapting pretrained Stable Diffusion v1.4, replacing text prompts with SigLIP features injected through trainable Adapter layers for garment reconstruction from clothed-person images. While effective as a baseline, TryOffDiff remains challenged by garments with complex geometry and structured patterns, largely due to its single-stream encoding path and sparse semantic tokens: compressing SigLIP features into 77 tokens emphasizes high-level semantics and inevitably attenuates dense, high-frequency texture cues, which degrades fine-grained spatial recovery and limits deformation inference. Motivated by this gap, our AlignVTOFF bypasses the standard downsampling bottleneck and preserves dense spatial features for high-fidelity canonical garment reconstruction.

\subsection{Latent Diffusion Models}
Latent Diffusion Models (LDMs)~\cite{rombach2022high} have become a dominant paradigm for high-quality generation and editing due to their strong pretrained priors and scalable conditioning interfaces. They have been widely adopted for text-to-image generation~\cite{esser2024scalingrectifiedflowtransformers} and for conditional synthesis tasks such as super-resolution~\cite{saharia2021imagesuperresolutioniterativerefinement}, pose-guided generation~\cite{shen2024advancing,shen2024imagpose}, and story visualization~\cite{shen2025boosting}. A key advantage of LDMs is extensibility: conditional control modules such as ControlNet~\cite{zhang2023adding} can be attached to improve fine-grained controllability beyond natural language, enabling constraints on layout, edges, pose, or other structural signals. In parallel, image-conditioned adapters such as IP-Adapter~\cite{ye2023ipadaptertextcompatibleimage} guide generation by injecting reference-image semantics into the diffusion denoiser, bridging text and visual conditions in a unified framework.

This line of work also motivates diffusion-based fashion synthesis and dressing systems. Notably, IMAGDressing~\cite{shen2025imagdressing} proposes a parallel U-Net design that separately captures garment appearance cues and structural guidance, and integrates multi-modal conditions (garment features, pose, and text prompts) through a hybrid attention mechanism during denoising. This parallel-stream strategy highlights that decoupling appearance from structure can be crucial for high-quality synthesis. However, VTOFF is more demanding than Virtual Dressing in terms of spatial reconstruction and texture restoration, because it requires recovering a canonical flat-lay garment from a real clothed-person photo rather than synthesizing a dressed person image under controllable conditions. Therefore, AlignVTOFF also adopts a parallel U-Net framework tailored to VTOFF, explicitly separating texture preservation from structural recovery. In particular, we inject dense, high-frequency texture cues through Texture-Spatial Feature Alignment (TSFA) to compensate for the information loss incurred by standard downsampling, enabling clearer and more faithful canonical garment reconstruction than single-branch LDM baselines.

\begin{figure}
    \centering
    \includegraphics[width=1\linewidth]{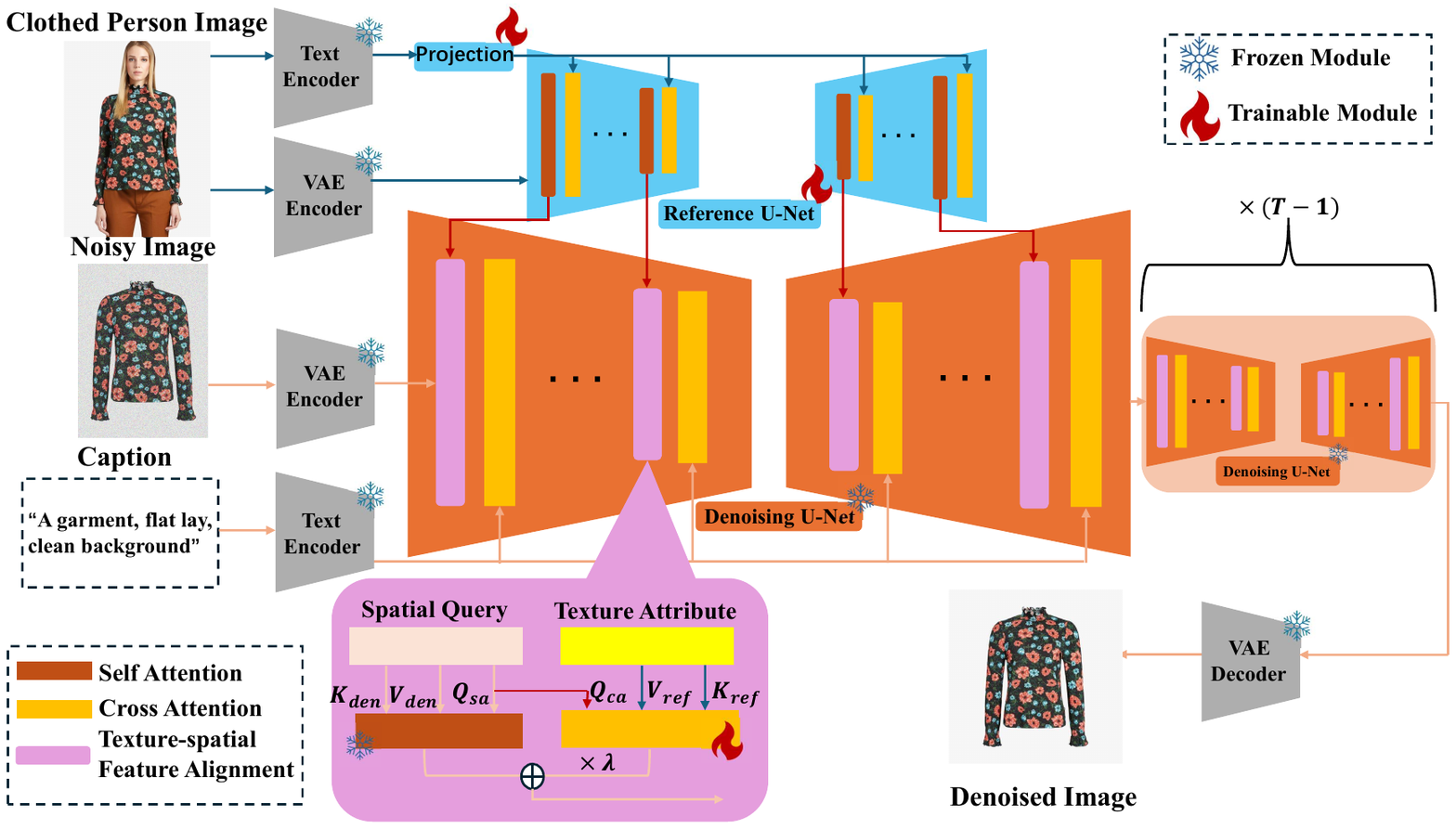}
\caption{Overview of the AlignVTOFF framework. It mainly consist of a trainable Reference U-Net and a frozen Denoising U-Net. The Reference U-Net extract features from the clothed-person image and inject it into the Denoising U-Net via TSFA.}
\label{fig:2}
\end{figure}

\section{Proposed Method}\label{sec:method} 
As shown in Fig.\ref{fig:2}, AlignVTOFF applies a parallel Stable Diffusion~\cite{rombach2022high} (SD v1.5) framework, a latent diffusion model designed for text-conditioned image generation. And AlignVTOFF mainly consists of a Reference U-Net and a Denoising U-Net. In the upper part, the trainable Reference U-Net is mainly used for feature extraction, but the difference lies in the Reference U-Net’s only ability to capture texture features from VAE and clothed person semantic features from CLIP. The lower part is the frozen Denoising U-Net adapts the SD v1.5 to generate garment images, we use a trainable TSFA to replace the original Self-Attention Modules.Therefore, we can inject features from clothed-person images into the Denoising U-Net. In addition, AlignVTOFF includes an image encoder and projection layer for encoding clothed-person image features, as well as a text encoder for encoding textual features.
\subsection{Reference U-Net}  
The extraction of fine-grained features, such as texture and structural edges, is essential for preserving the fidelity of details within VTOFF tasks. As mentioned, the trainable Reference U-Net capture texture features from VAE and clothed person semantic features from CLIP. Specifically, given a clothed-person image $X \in \mathbb{R}^{3 \times H \times W}$, we first convert it into a latent space representation $Z_g \in \mathbb{R}^{4 \times H/8 \times W/8}$ using a frozen VAE Encoder. Simultaneously, using a frozen CLIP image encoder from IP-Adapter and a trainable projection layer to extract token embeddings from $X$. Furthermore, these features interact in the cross attention modules. In addition, it has to be mentioned that Reference U-Net is only used for image encoding. Therefore, no noise adding into the clothed-person images, only a single forward pass is performed during the process.
\subsection{Denoising U-Net}  
The lower part of the framework is the frozen Denoising U-Net, which perform the denoising process in the latent space. Formally, the VAE utilizes an encoder $\mathcal{E}$ to compress the input image $\mathbf{x}$ into a latent representation $\mathbf{z} = \mathcal{E}(\mathbf{x})$. Conversely, the VAE decoder $\mathcal{D}$ reconstructs the image $\mathbf{x}$ from the latent representation $\mathbf{z}$, \textit{i.e.}, $\mathbf{x} = \mathcal{D}(\mathbf{z})$. With the unique objective of Virtual Try-Off (VTOFF) tasks, which is to generate flat-lay garment images, the input of CLIP text encoder remains the prompt "A garment, flat lay, clean background". The CLIP text encoder transforms this prompt into token embeddings $\mathbf{c}$. Additionally, the original cross attention modules are replaced by the proposed TSFA, which are detailed in the subsequent section.
During the process, Gaussian noise $\epsilon$ is added to the latent representation $\mathbf{z}$ by time step $t$ to produce $\mathbf{z}_t$, where $t \in [0, T]$. 
Subsequently, the Denoising U-Net estimates the noise residual within $\mathbf{z}_t$. To optimize the network parameters $\theta$, for each timestep $t$, the model is trained by a mean squared error objective $\mathcal{L}_{\text{LDM}}$:
\begin{equation}
    \mathcal{L}_{\text{LDM}} = \mathbb{E}_{\mathbf{z}_t, \epsilon \sim \mathcal{N}(0, I), \mathbf{c}, t} \left[ \| \epsilon_\theta(\mathbf{z}_t, \mathbf{c}, t) - \epsilon \|^2 \right],
    \label{eq:ldm_loss}
\end{equation}
where $\mathbf{z}_t = \sqrt{\alpha_t}\mathbf{z}_0 + \sqrt{1-\alpha_t}\epsilon_t$ and $\epsilon_t$ is the sampled noise at timestep $t$. $\mathbf{z}_0$ is the latent representation derived from the real flat-lay garment image $\mathbf{x}$ via the VAE encoder $\mathcal{E}$ \textit{i.e.}, $\mathbf{z}_0 = \mathcal{E}(\mathbf{x}_0)$, conditioned on the text prompt $\mathbf{c}$.
On the sampling stage, the final noise prediction is synthesized by the conditional model $\epsilon_\theta(\mathbf{x}_t, \mathbf{c}, t)$ and
the unconditional model $\epsilon_\theta(\mathbf{x}_t,\emptyset,t)$ via classifier-free guidance~\cite{ho2022classifier}:
\begin{equation}
    \hat{\epsilon}_\theta(\mathbf{x}_t, \mathbf{c}, t) = w\epsilon_\theta(\mathbf{x}_t, \mathbf{c}, t) + (1-w)\epsilon_\theta(\mathbf{x}_t,\emptyset,t).
    \label{eq:cfg}
\end{equation}
where $w$ is the guidance scale, deciding the influence of the condition $\mathbf{c}$.

In addition,to ensure the extensibility of AlignVTOFF, we maintain the Denoising U-Net in a frozen state. This allows our framework to function as a plug-and-play system. Specifically, structural priors from ControlNet~\cite{zhang2023adding} are injected into the Denoising U-Net's encoder via zero-convolutions, while the TSFA simultaneously incorporates fine-grained texture features from the Reference U-Net without conflict.
\subsection{Texture-Spatial Feature Alignment}
For VTOFF tasks, the Denoising U-Net, except for capability to generate images, it should also integrate features that extracted from the clothed-person image, which is the function of this TSFA. As shown in Fig.~\ref{fig:2}, all original self-attention modules are replaced by TSFA. Consisting of a frozen self-attention module and a trainable cross-attention module, the output of TSFA $\mathbf{Z}_{tsfa}$ combines the spatial features $\mathbf{Z}_{den}$ from the Denoising U-Net and the reference features $\mathbf{C}_{ref}$ from the Reference U-Net:
\begin{equation}
    \mathbf{Z}_{tsfa} = \underbrace{\text{Softmax}\left(\frac{Q_{sa} K_{den}^\top}{\sqrt{d}}\right) V_{den}}_{\text{Self-Attention}} + \lambda \underbrace{\text{Softmax}\left(\frac{Q_{ca} K_{ref}^\top}{\sqrt{d}}\right) V_{ref}}_{\text{Cross-Attention}}.
\end{equation}
where the coefficient $\lambda \in [0, 1.5]$ is a hyperparameter deciding the influence of the $\mathbf{C}_{ref}$. The projection matrices are defined as $Q_{sa}=\mathbf{Z}_{den} W_q$, $K_{den}=\mathbf{Z}_{den} W_k$, $V_{den}=\mathbf{Z}_{den} W_v$, $K_{ref} =\mathbf{C}_{ref} W'_k$, and $V_{ref} = \mathbf{C}_{ref} W'_v$. Crucially, as indicated by the red arrow in Fig.~\ref{fig:2}, we implement a shared-query mechanism where the spatial query is reused, i.e., $Q_{ca} = Q_{sa}$. To maintain the generating capability of the pre-trained model, the self-attention module is frozen, which means only the weights $W'_k$ and $W'_v$ of the cross attention module are trainable. This selective fine-tuning strategy allows our model to learn the clothed-person images while retaining the robust generating capability of the original SD v1.5. The architectural design of TSFA is rooted in the necessity of balancing the generative stability of the pre-trained model with the fidelity of external texture injection. Unlike basic attention blocks that rely solely on a single attention mechanism, our TSFA is designed to mitigate the trade-off between identity preservation and structural integrity. If the TSFA were configured exclusively with self-attention, the Denoising U-Net would struggle to perceive the complex texture cues from the Reference U-Net, ultimately leading to a significant loss of garment identity. On the other hand, a purely cross-attention-based configuration would force the denoising process to over-attend to reference features, which often disrupts the spatial continuity and generative priors inherent in the pre-trained Stable Diffusion model, resulting in structural distortions or artifacts.

\subsection{Loss Function}  
While the $\mathcal{L}_{\text{LDM}}$ loss effectively drives the U-Net's denoising capability, it still struggles to capture high-frequency textures and human perceptual discrepancy in detail-rich image generation tasks, such as VTOFF tasks. Therefore, to realize the visual realism and assure fidelity of the generated image, we introduce the perceptual loss $\mathcal{L}_{\text{LPIPS}}$~\cite{zhang2018unreasonable} into our overall training loss functions.
The Learned Perceptual Image Patch Similarity ($\mathcal{L}_{\text{LPIPS}}$) leverages features from a pre-trained deep feature extractor (e.g., VGG~\cite{simonyan2014very}) to measure the difference between the real image $\mathbf{x}$ and the generated image $\mathbf{\hat{x}}$. Minimizing this loss, we encourage the model to generate garment images with perceptually consistent details. In summary, our training objective $\mathcal{L}_{\text{total}}$ is a weighted sum of the LDM loss $\mathcal{L}_{\text{LDM}}$ and the perceptual loss 
$\mathcal{L}_{\text{LPIPS}}$:
\begin{equation}
\mathcal{L}_{\text{total}} = \mathcal{L}_{\text{LDM}} + \lambda_{\text{LPIPS}} \cdot \mathcal{L}_{\text{LPIPS}}.
\end{equation}
Empirically, we set the coefficient $\lambda$ to 0.05 to keep the balance between perceptual texture quality and pixel-level reconstruction fidelity. This ensures that the model focuses on structural consistency via the MSE term the LPIPS term to refine high-frequency details without instability of training.

\subsection{Comparison with Existing Architectures}
To further clarify the advance behind AlignVTOFF, we highlight the key differences between our approach and existing reference-based diffusion models. First, compared with general-purpose adapters such as IP-Adapter~\cite{ye2023ipadaptertextcompatibleimage}, which often suffer from spatial misalignment when handling complex garment textures, our TSFA module employs a shared-query mechanism. This ensures that the denoising process maintains the structural integrity of the pre-trained model while precisely injecting fine-grained features from the Reference U-Net. Second, traditional fine-tuning methods often compromise the high-quality generation capabilities of Stable Diffusion (SD). By keeping the Denoising U-Net frozen and only optimizing the TSFA and projection layers, AlignVTOFF acts as a plug-and-play framework. This allows for seamless integration with other structural guidance tools without retraining the core architecture, a flexibility that many end-to-end VTOFF models lack. Furthermore, While some models rely solely on CLIP embeddings, which may lose local texture details, AlignVTOFF extracts both latent texture features from the VAE and semantic guidance from CLIP. This hybrid approach addresses the "identity loss" problem commonly found in VTOFF tasks.

\begin{figure}
	\centering
	\includegraphics[width=0.8\linewidth,trim=0 100 0 120, clip]{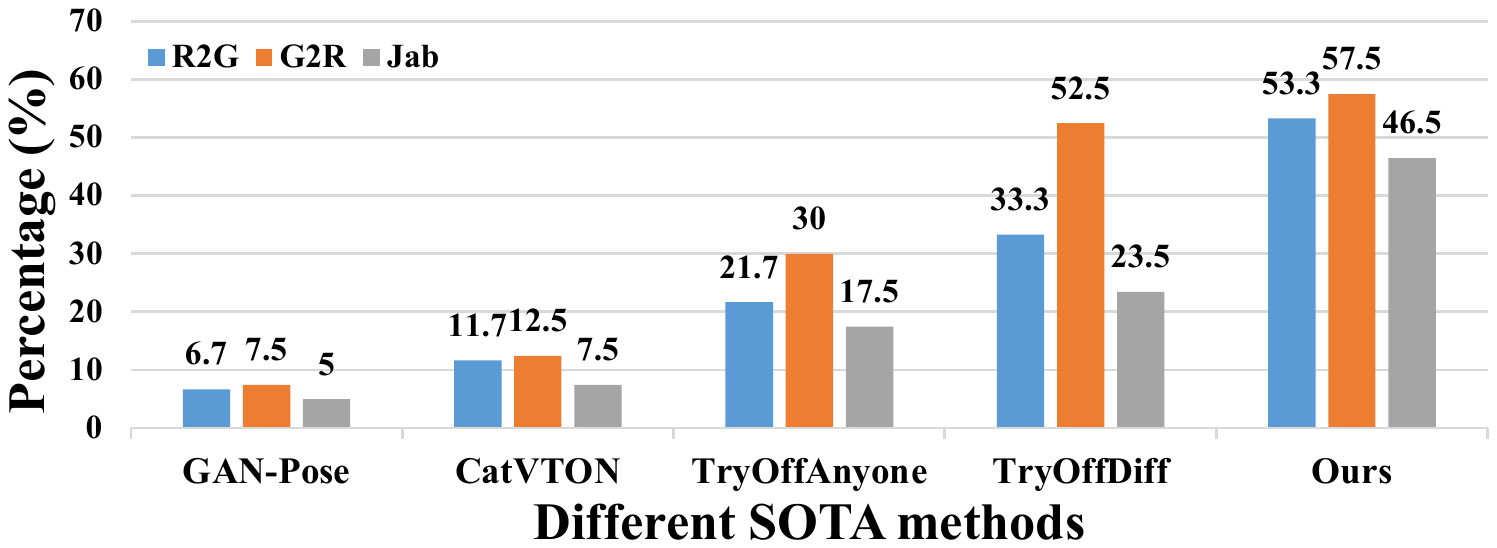}
	\caption{User study results on VITON-HD in terms of R2G, G2R and Jab metric. Higher values in these three metrics indicate better performance.}
	\label{fig:3}
\end{figure}

\section{Experiment and Analysis}\label{sec:exp}  

To validate the proposed Virtual Try-Off(VTOFF) method's superiority, it is compared with multiple state-of-the-art VTOFF approaches on two large-scale datasets, namely, VITON-HD~\cite{choi2021viton} and DressCode~\cite{morelli2022dress}.

\subsection{Datasets}

\subsubsection{\emph{VITON-HD}}
VITON-HD~\cite{choi2021viton} is a widely recognized benchmark for high-resolution virtual try-on tasks, featuring a vast collection of 13,679 pairs of half-body models and upper-body garments at a resolution of $1024\times 768$. The dataset is characterized by diverse garment textures, complex body poses, and high-fidelity details, providing a challenging environment for evaluating garment reconstruction. However, due to the problem of data leakage, we adopt the  resulting cleaned dataset in TryOffDiff~\cite{velioglu2025tryoffdiffvirtualtryoffhighfidelitygarment}, which contains 11,552 unique image pairs for training and 1,990 unique image pairs for testing.

\subsubsection{\emph{DressCode}}
DressCode~\cite{morelli2022dress}is a widely dataset consisting of  53,792 pairs ($1024\times 768$) of full-body models and upper-body, lower-body, and dress garments. Unlike VITON-HD, which primarily features half-body models, DressCode provides full-body images with more complex backgrounds and a wider variety of human poses. To maintain consistency with our focus on upper-body VTOFF tasks, we specifically extract the upper-body sub-category for our experiments. Following the official split, we adopt 13,563 training and 1,800 test upper-body pairs for Dress Code.

\subsection{Metrics}  
To evaluate the methods comprehensively, we adopt eight metrics across different dimensions: SSIM~\cite{wang2004image}, $\text{MS}^{\text{SSIM}}$~\cite{wang2003multiscale} and $\text{CW}^{\text{SSIM}}$~\cite{wang2005translation} for geometric and structural reconstruction accuracy. Specifically, $\text{CW}^{\text{SSIM}}$ provides robustness against small spatial translations inherent in the VTOFF process. While SSIM ensures pixel alignment, we prioritize LPIPS~\cite{zhang2018unreasonable} and DISTS~\cite{ding2020image} to evaluate textural fidelity and perceptual realism. These metrics are critical for validating the reconstruction of complex high-frequency patterns. FID~\cite{parmar2022aliased}, KID~\cite{binkowski2018demystifying} for global statistical similarity and image realism. These evaluation metrics ensure an objective assessment of both visual quality and noise suppression.

\subsection{Implementation Details} 
AlignVTOFF is based on SD v1.5, we initialize the weights of our Reference U-Net by inheriting the pre-trained weights of U-Net in SD v1.5 and then finetuning it. Input images ($1024\times 768$) from datasets are resized to resolution of $512\times 512$ for Reference U-Net. Divided into two stages, the training process is totally 53k steps on a single NVIDIA GPU with 48GB VRAM. In Stage 1 (50k steps), we focus on global structural alignment. We train the model with simple MSE loss.In this stage, We use AdamW optimizer to train parameters with a linearly increased learning rate from 0 to $1 \times 10^{-5}$ through the first 500-steps warm-up, which will be maintained later. In addition, the effective batch size is 16 ($4\times 4$) by using a batch size of 4 combined with gradient accumulation over 4 steps. In stage 2 (3k steps), we pivot to fine-grained perceptual refinement. With a combination of MSE and LPIPS loss with a weighting factor of $\lambda_\text{LPIPS} = 0.05$, we skip the warmup process and adopt a constant learning rate of $3 \times 10^{-6}$ with a batch size of 4. At the inference process, we generate the images with the DPM++ for 25 sampling steps and set the guidance scale to 2.0.

\subsection{Comparison with State-of-the-art Methods} 

\subsubsection{Comparisons on VITON-HD}
Quantitative results on VITON-HD are shown in Table~\ref{tab:1}, AlignVTOFF outperforms existing state-of-the-art methods across the majority of metrics. Specifically, in terms of image realism and perceptual fidelity, our model achieves a DISTS of 21.6 and an LPIPS of 24.7, significantly surpassing the second-best method, TryOffDiff. This superiority is attributed to the proposed TSFA, which injects the spatial features into the Denoising U-Net. By establishing a dense spatial correspondence, the TSFA effectively bridges the gap between global semantic understanding and local textural reconstruction, thereby mitigating the information loss typically associated with sparse token-based alignment. Our framework effectively preserves the structural integrity of complex patterns that are typically lost in lightweight encoding modules. 

\begin{table}[t] 
\centering
\renewcommand{\arraystretch}{1.3} 
\setlength{\tabcolsep}{2pt}
\caption{Quantitative comparison on the VITON-HD dataset. Our proposed AlignVTOFF achieves a superior balance between structural alignment and perceptual realism.}
    \begin{tabular}{l|ccc|cccc}
    \toprule
    \rowcolor{gray!10} 
    Method & SSIM $\uparrow$ & $\text{MS}^{\text{SSIM}}$ $\uparrow$ & $\text{CW}^{\text{SSIM}}$ $\uparrow$ &DISTS $\downarrow$ & LPIPS $\downarrow$ & FID $\downarrow$ & KID $\downarrow$\\
    \midrule
    TryOffDiff~\cite{velioglu2025tryoffdiffvirtualtryoffhighfidelitygarment} & \underline{77.9} & \underline{69.1} & 49.2&\underline{22.0} & \underline{32.8} & 18.1 & 5.7 \\
    CatVTON~\cite{chong2024catvton}   & 72.8 & 56.9 & 32.0 & 28.2 & 45.9 & 31.4 & 17.8 \\
    GAN-Pose~\cite{roy2023multi} & 77.4 & 63.8 & 32.5 &30.4 & 44.2 & 73.2 & 55.8 \\
    TryOffAnyone~\cite{xarchakos2024tryoffanyone}& 75.9 & 58.3 & \underline{51.3} & 23.5 & 35.2 & \textbf{12.7} & \textbf{2.9} \\ 
    \midrule
    \textbf{AlignVTOFF (Ours)} & \textbf{78.1} & \textbf{70.1} &\textbf{52.5} &\textbf{21.6} & \textbf{24.7} & \underline{14.7} & \underline{4.4} \\
    \bottomrule
    \end{tabular}
\label{tab:1}
\end{table}

\begin{table}[t] 
\centering
\renewcommand{\arraystretch}{1.3} 
\setlength{\tabcolsep}{2pt}
\caption{Quantitative comparison on the DressCode dataset (upper-body subset). Our proposed AlignVTOFF achieves a superior balance between structural alignment and perceptual realism.}
    \begin{tabular}{l|c|cccc}
    \toprule
    \rowcolor{gray!10} 
    Method & SSIM $\uparrow$ & DISTS $\downarrow$ & LPIPS $\downarrow$ & FID $\downarrow$ & KID $\downarrow$\\
    \midrule
    TryOffDiff~\cite{velioglu2025tryoffdiffvirtualtryoffhighfidelitygarment}& \underline{76.6} & 29.0 & 40.6 & 38.0 & 17.3 \\
    Any2anytryon~\cite{guo2025any2anytryon}   &\underline{76.6} & \underline{25.8} & \underline{39.0} & \textbf{15.8} & \textbf{3.2} \\
    \midrule
    \textbf{AlignVTOFF (Ours)} & \textbf{80.2} & \textbf{23.5} & \textbf{27.0} & \underline{16.9} & \underline{4.2} \\
    \bottomrule
    \end{tabular}

\label{tab:2}
\end{table}

\subsubsection{Comparisons on DressCode}
The quantitative results on the DressCode dataset are presented in Table~\ref{tab:2}. Our evaluation focuses specifically on the upper-body subset to ensure a consistent comparison. As shown in the table, AlignVTOFF consistently achieves the best performance across the majority of metrics. Specifically, our model reaches a DISTS of 23.5 and an LPIPS of 27.0, outperforming other methods. This performance gain indicates that our TSFA module effectively captures the complex textures and details inherent in diverse upper-body styles, such as varying necklines and sleeve patterns. By leveraging a frozen generative backbone, our framework successfully preserves high-frequency details while ensuring the synthesis of high-fidelity, standardized flat-lay images.

\begin{figure}
    \centering
    \includegraphics[width=1\linewidth,trim=0 80 0 100, clip]{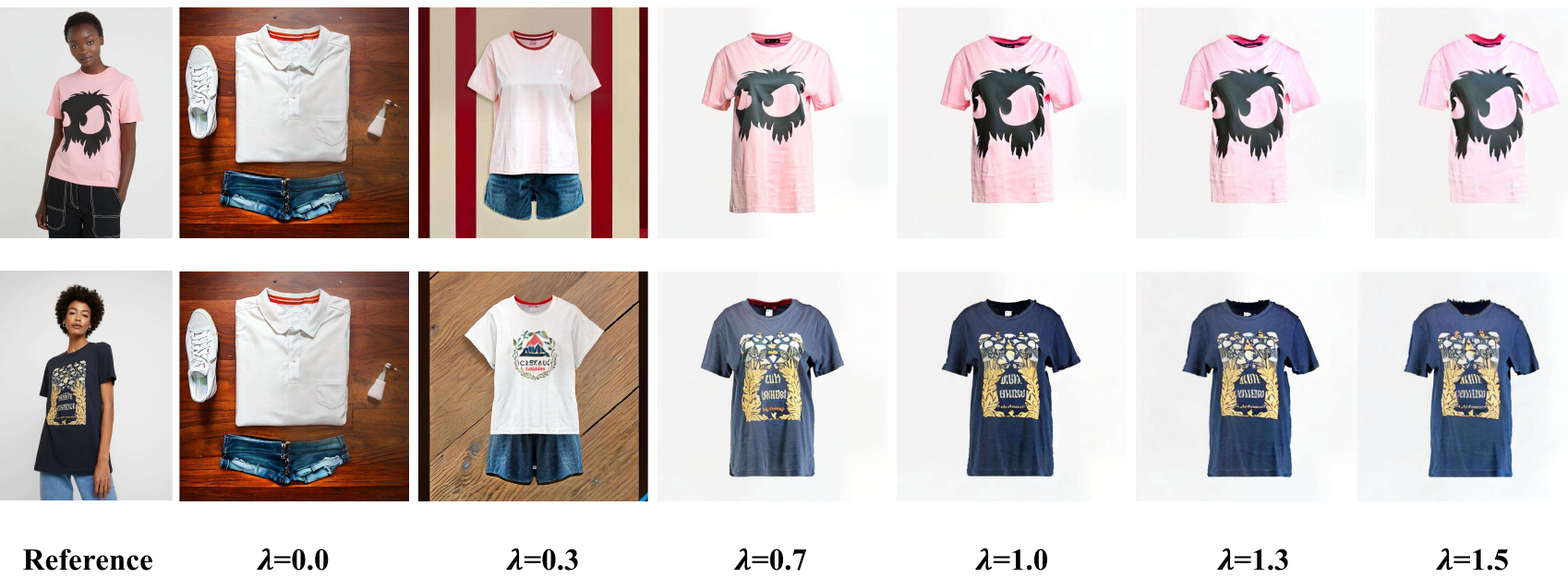}
\caption{Qualitative comparison on results with different coefficient $\lambda$.}
\label{fig:4}
\end{figure}

\subsection{User Study}
The aforementioned quantitative and qualitative comparisons demonstrate the clear superiority of our proposed AlignVTOFF. However, since VTOFF task is inherently centered on human perception, we conducted a user study involving 20 participants with expertise in computer vision to further validate the subjective quality of our results. This study included comparisons with fundamental facts (i.e., R2G and G2R) and comparisons with other methods (i.e., Jab), where higher scores across these three metrics signify superior performance. As illustrated in Fig.~\ref{fig:3}, AlignVTOFF consistently achieves leading results on the VITON-HD dataset. Notably, in terms of the G2R metric, \qty{57.5}{\percent} of our generated images were perceived as authentic—outperforming the baseline by a significant margin. Furthermore, our Jab score of \qty{46.5}{\percent} reflects a clear participant preference for our approach. This significant advantage in subjective preference is primarily attributed to the TSFA module's precision in detail restoration; participants noted that AlignVTOFF excels in maintaining garment logo integrity and texture clarity, effectively mitigating the "texture blurring" artifacts common in conventional diffusion models.

\subsection{Ablation Studies and Analysis}  
The comparison results presented in Table~\ref{tab:1}, Table~\ref{tab:2}, demonstrate that the proposed VTOFF method is superior to many state-of-the-art VTOFF methods.
In what follows,  the proposed VTOFF method is comprehensively analyzed from three key aspects to investigate the logic behind its superiority.

\subsubsection{Role of Each Component} 
\begin{table}[t]
\centering
\renewcommand{\arraystretch}{1.3} 
\caption{ Quantitative results for different components on VITON-HD. IEB, TSFA and CA
denote the image encoder branch, TSFA and cross attention.}
    \begin{tabular}{ccc|c|cc}
    \toprule
    \rowcolor{gray!10}
Case & IEB & TSFA & SSIM $\uparrow$ & DISTS $\downarrow$ & LPIPS $\downarrow$\\ \midrule
1    &          &          & 36.6              &   50.7           & 81.2              \\
2    & \checkmark &    replaced by CA     &   76.8          &    23.7          &     28.9         \\
3    & \checkmark &    encoder-only     &    63.4         &     33.4        &    46.0       \\
4    & \checkmark &    decoder-only     &     75.7        &       24.0        &      31.5        \\
\midrule
5    & \checkmark & \checkmark & \textbf{78.1}     & \textbf{21.6}     & \textbf{24.7}     \\ 
\bottomrule
\end{tabular}
\label{tab:3}
\end{table}

As shown in Table.\ref{tab:3}, an ablation study evaluates the individual contributions of the proposed Image Encoder Branch (IEB) and Texture-Spatial Feature Alignment (TSFA). Compared to the baseline without IEB and TSFA (Case 1), the inclusion of IEB (Case 2) has significant improvements across all metrics, with SSIM improving from 36.6 to 76.8, validating its efficacy in capturing semantic features. The integration of TSFA (Case 5), which combines IEB and TSFA, achieves the superior performance, further boosting quantitative results. This demonstrates that while IEB establishes the global context, TSFA provides essential spatial-wise guidance, which is important for high-fidelity garment reconstruction and precise texture alignment.

\subsubsection{Influence of TSFA Mechanism} 
As shown in Table~\ref{tab:3}, we investigate the layer-wise contribution of the TSFA mechanism by selectively applying it to the encoder (Case 3) and decoder (Case 4) stages of the Denoising U-Net.Specifically, the full TSFA integration (Case 5) achieves the best performance, yielding a peak SSIM of 78.1 and a lower LPIPS of 24.7. The encoder-only configuration effectively establishes the global layout and structural skeleton, yet it fails to propagate sufficient high-frequency textural information to the final synthesis layers, resulting in blurred garment appearances. Conversely, while the decoder-only configuration succeeds in recovering intricate surface details, the absence of early-stage structural guidance leads to severe spatial distortion and latent misalignment between the garment and the target grid. This contrast suggests that the encoder acts as a spatial anchor that stabilizes the geometric proportions, while the decoder serves as a high-resolution refiner for texture mapping. Consequently, the seamless integration of encoder-side spatial anchoring and decoder-side texture rendering is indispensable for harmonizing global geometric integrity with local textural fidelity in high-fidelity garment generation.

\begin{figure}
    \centering
    \includegraphics[width=0.6\linewidth,trim=60 60 60 60, clip]{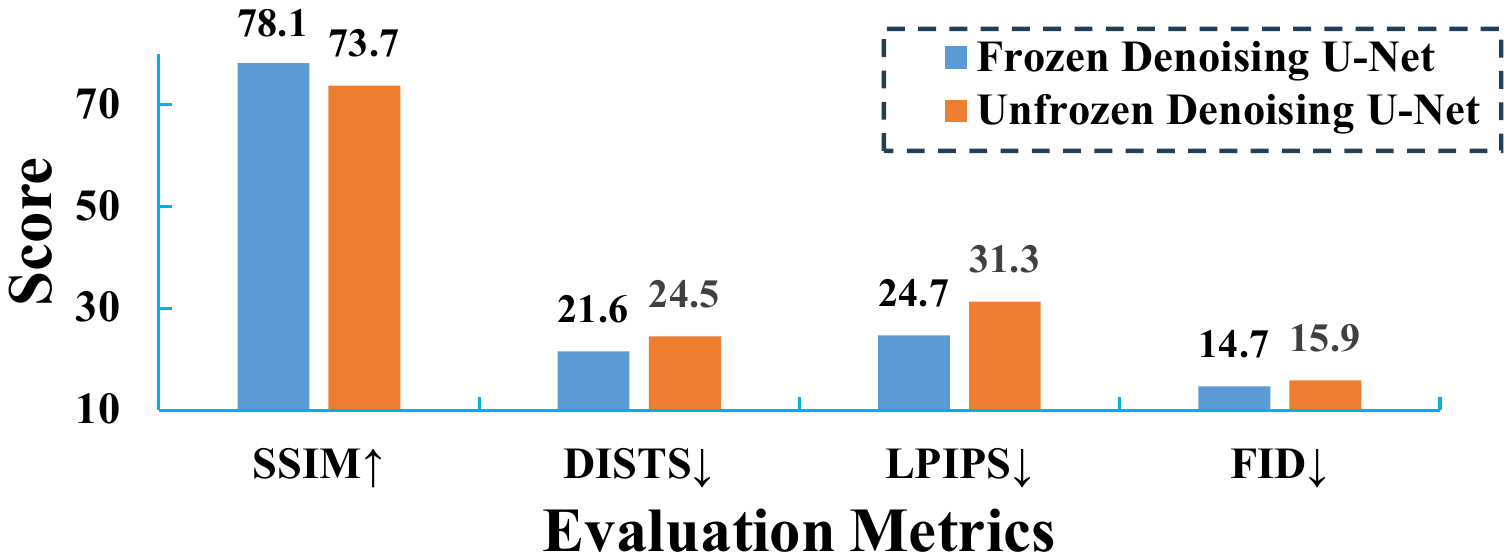}
\caption{Quantitative analysis of the Denoising U-Net freezing strategy on VITON-HD.}
\label{fig:5}
\end{figure}
\subsubsection{Impact of Denoising U-Net Freezing} 
To further explore the optimal learning strategy, we investigate the impact of unfreezing the pre-trained Denoising U-Net weights. Quantitative results shown in Fig.~\ref{fig:5}, maintaining a frozen Denoising U-Net consistently outperforms its unfrozen variant. Specifically, the frozen configuration achieves a lower DISTS of 21.6 and a lower LPIPS of 24.7. This suggests that the extensive generative priors embedded in the frozen weights are crucial for maintaining image quality, whereas unfreezing these layers can lead to the disturbance of pre-trained knowledge and potential overfitting to specific training-set distributions. Specifically, unfreezing the backbone often triggers "catastrophic forgetting" of high-level aesthetic priors, where the model begins to produce structural artifacts by over-optimizing for the specific spatial noise and lighting patterns of the training data. Furthermore, this frozen constraint serves as a powerful structural regularizer, compelling the trainable Reference U-Net to adapt its feature maps to a stable, well-defined latent space and thereby ensuring superior zero-shot generalization to unseen garment styles. This reinforces the design choice of keeping the Denoising U-Net frozen to ensure both robustness and high-fidelity output. 

\begin{figure}
    \centering
    \includegraphics[width=1\linewidth]{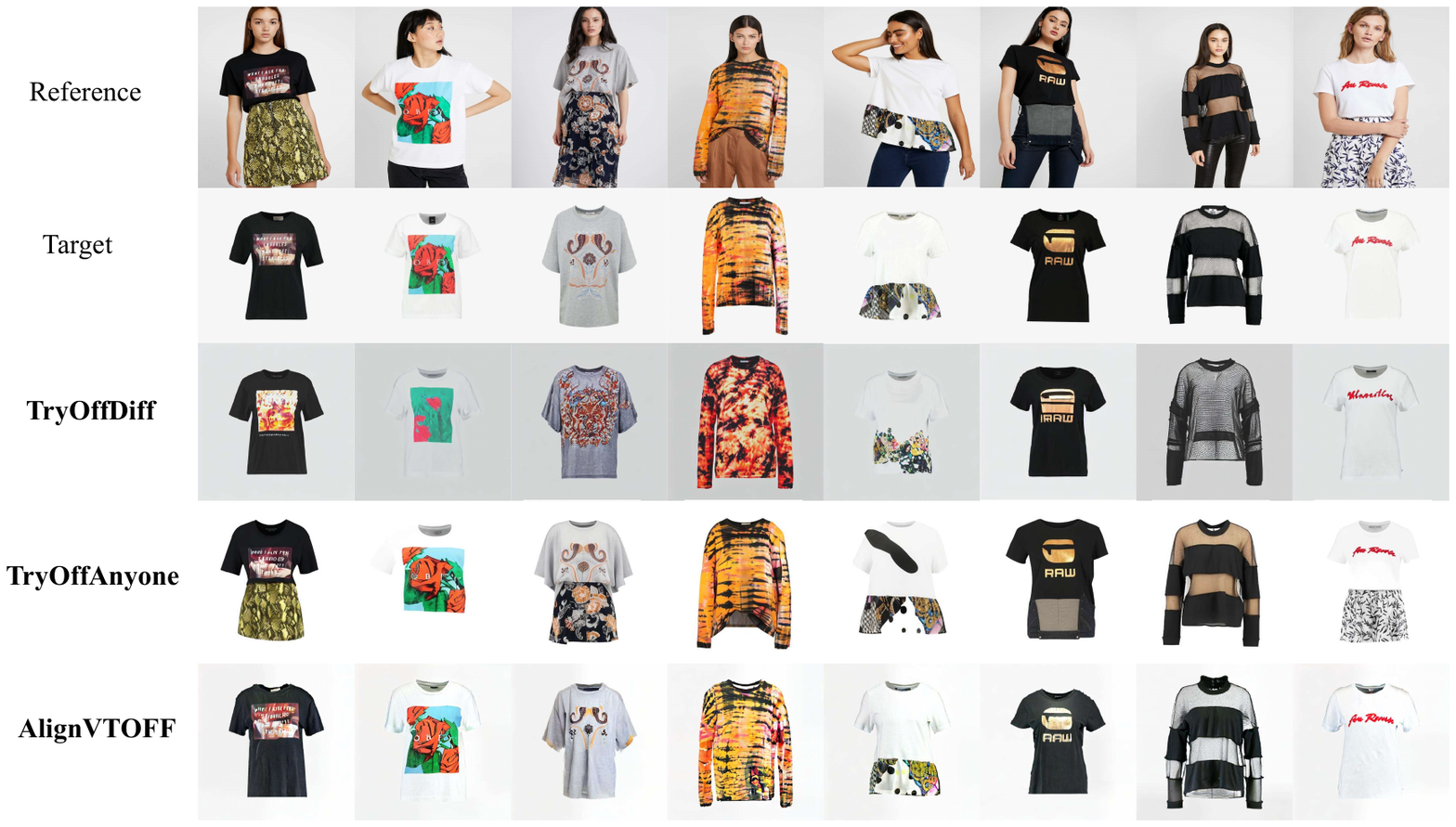}
\caption{Qualitative comparison on VITON-HD. Compared to TryOffDiff~\cite{velioglu2025tryoffdiffvirtualtryoffhighfidelitygarment} and TryOffAnyone~\cite{xarchakos2024tryoffanyone}, AlignVTOFF is capable of generating garment images with accurate structural details as well as fine textural details.}
\label{fig:6}
\end{figure}

\subsubsection{Potential Application}  
Due to the frozen-backbone of our methods, AlignVTOFF offers significant extensibility as a plug-and-play framework. It can combine with existing adapters, such as ControlNet~\cite{zhang2023adding}, during the inference stage. Specifically, our architecture enables a decoupled control mechanism where we can inject structural priors into the Denoising U-Net's encoder via zero-convolutions. Simultaneously, our proposed TSFA module precisely map textural details without interfering with the geometric guidance provided by external adapters. This synergistic combination suggests a feasible pathway where our TSFA modules ensure textural fidelity while auxiliary models manage geometry and style. This modularity demonstrates that AlignVTOFF is not just a standalone model, but a flexible model capable of high-precision, multi-conditioned garment generation.

\subsubsection{Sensitivity Analysis} 

Fig.~\ref{fig:4} shows the influence of the hyper-parameter $\lambda$ on generated samples using a fixed random seed. As $\lambda$ approaches 1.0, the synthesized garments exhibit increasing fidelity to the input reference. In addition, a smaller $\lambda$ prioritizes alignment with text prompts. Consequently, this case leads to a diminished correlation between the reconstructed garment and the reference image. Conversely, a larger value mainly focuses on preserving garment features, causing the sharp edges. Specifically, an excessive $\lambda$ compels the denoising process to over-attend to the dense spatial cues from the Reference U-Net, potentially overriding the generative model's inherent ability to produce smooth, natural fabric folds. This trade-off suggests that $\lambda$ effectively balances text-driven generation with reference-based editing. Based on this experiment, we empirically set $\lambda = 1.0$ for all subsequent experiments.

\begin{figure}
    \centering
    \includegraphics[width=0.7\linewidth,trim=50 80 50 60, clip]{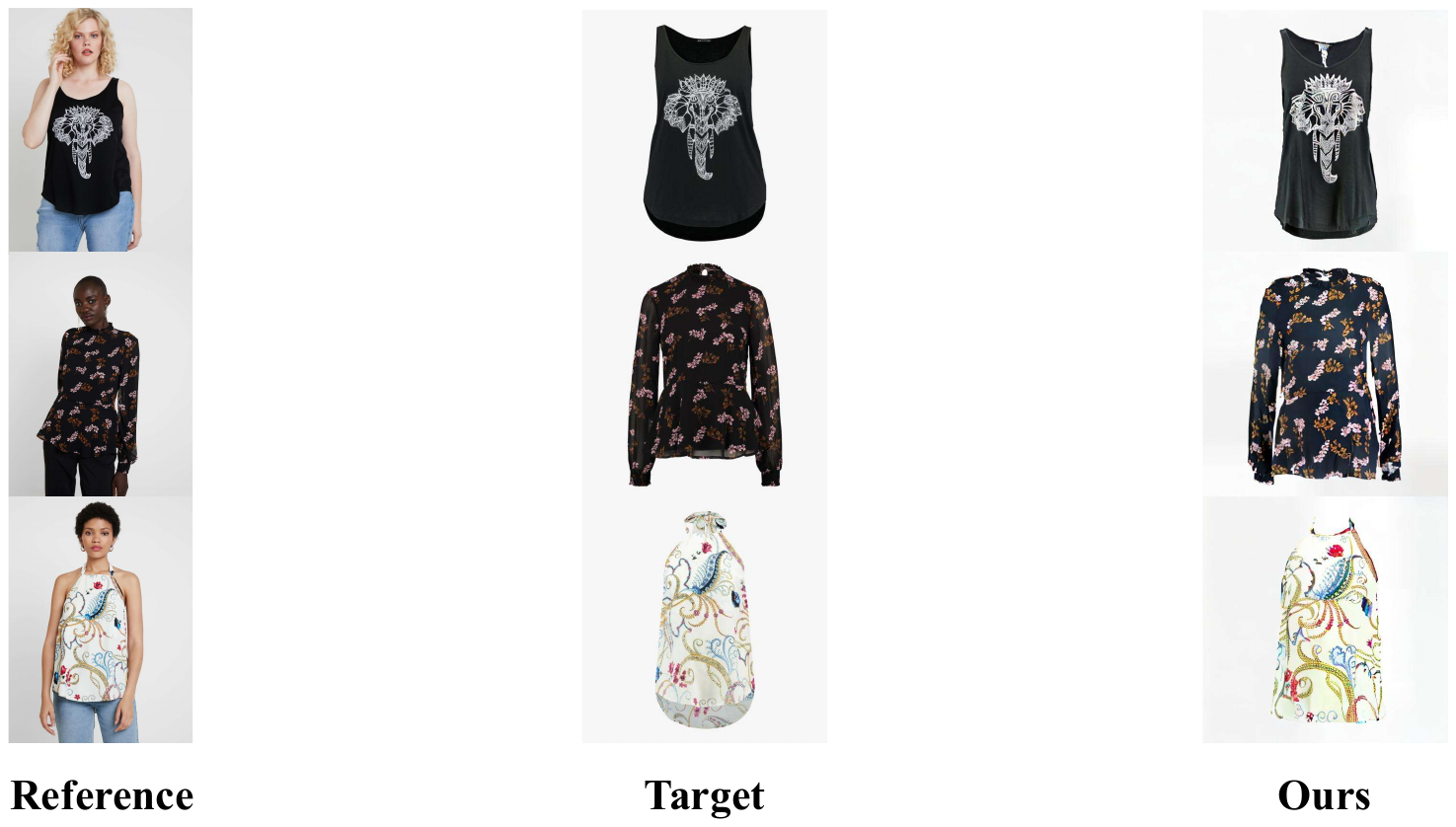}
\caption{Typical failure cases of AlignVTOFF.}
\label{fig:7}
\end{figure}
\subsection{Visualization}  
As shown in Fig.~\ref{fig:6}, compared to TryOffDiff~\cite{velioglu2025tryoffdiffvirtualtryoffhighfidelitygarment} and TryOffAnyone~\cite{xarchakos2024tryoffanyone}, our proposed AlignVTOFF has a superior ability to reconstruct complex high-frequency patterns while preserving the fundamental structural characteristics of the generated upper garment that closely align with the Ground Truth. This visual superiority is driven by the TSFA module, which facilitates the precise injection of multi-scale texture features into the Denoising U-Net. By leveraging the Parallel U-Net architecture, our framework effectively decouples texture mapping from geometric guidance, thereby preventing the loss of complex details that often occur in single-stream encoders. Specifically, while conventional single-stream models often collapse spatial information into sparse semantic tokens, our TSFA mechanism establishes a dense spatial correspondence that anchors high-frequency textures precisely to their target latent coordinates. This parallel-stream design effectively bypasses the information bottleneck inherent in standard downsampling paths, ensuring that intricate structured patterns are reconstructed with superior clarity rather than being smoothed out by the low-pass effect of semantic-heavy encoding. Ultimately, these qualitative improvements directly prove the significant performance gains observed in our quantitative evaluation metrics, such as DISTS and LPIPS.

\subsection{Failure Cases and Limitation}
Despite the superior performance, AlignVTOFF still exhibits limitations in chromatic reconstruction. As illustrated in Fig.~\ref{fig:7}, a primary issue is the chromaticity discrepancy, where the generated garment colors appear slightly brighter than the ground truth. This can be attributed to the inherent bias of the frozen Denoising U-Net's generative priors, which may prioritize image cleanliness over color intensity. In addition, the normalization process within the Reference U-Net during feature extraction might inadvertently act as a low-pass filter on color information, smoothing out high-saturation peaks in the feature space before they reach the TSFA module. Furthermore, the systematic reduction of these high-frequency spectral components during the alignment phase implies that the model may interpret extreme saturation as potential noise, thereby prioritizing a clean reconstruction over absolute chromatic accuracy. Future research will explore color-aware loss functions to further enhance the fidelity of chromatic reconstruction.

\section{Conclusion}\label{sec:con} 
VTOFF was identified as a challenging multimodal image generation problem. It was difficult to preserve features due to complex high-frequency details and geometric deformation. Existing VTOFF methods focused on generating images through lightweight modules, which had limited capability to generate garments with complex structured patterns. Our proposed AlignVTOFF model tended to preserve garment complex details and reconstructed them via a Reference U-Net and TSFA. The Reference U-Net is designed to address the difficulty of geometric distortion and complex pattern detail extraction by multi-scale feature extraction explicit geometric fidelity enhancement. In addition, to integrates the refined garment features extracted, we present the TSFA, consisting of a trainable cross attention module and a frozen self-attention module. Extensive experiments demonstrate that our AlignVTOFF achieves state-of-the-art performance under various controlled conditions. Further work will focus on current limitations, extending the AlignVTOFF architecture to handle color-aware loss and diverse garments, further advancing the applicability of VTOFF technology.

\section*{Declaration of generative AI and AI-assisted technologies in the manuscript preparation process}

During the preparation of this work the author(s) used Gemini in order to refine the academic language, improve the structural organization of the manuscript, and ensure compliance with formatting guidelines (e.g., refining research highlights). After using this tool/service, the author(s) reviewed and edited the content as needed and take(s) full responsibility for the content of the published article.

\section*{Funding Sources}
This research did not receive any specific grant from funding agencies in the public, commercial, or not-for-profit sectors.

\section*{Acknowledgments}
The authors would like to express their gratitude to AutoDL (www.autodl.com) for providing the high-performance GPU computing clusters that supported the numerical simulations and model training in this research. The authors also thank the National Demonstration Center for Experimental Engineering Training Education , Shanghai University for the institutional support.

\section*{Declaration of Competing Interest}
The authors declare that they have no known competing financial interests or personal relationships that could have appeared to influence the work reported in this paper.

\bibliographystyle{cas-model2-names}

\bibliography{cas-refs}

@article{shen2025imagedit,
  title={IMAGEdit: Let Any Subject Transform},
  author={Shen, Fei and Xu, Weihao and Yan, Rui and Zhang, Dong and Shu, Xiangbo and Tang, Jinhui},
  journal={arXiv preprint arXiv:2510.01186},
  year={2025}
}

@article{shen2025imagharmony,
  title={IMAGHarmony: Controllable Image Editing with Consistent Object Quantity and Layout},
  author={Shen, Fei and Du, Xiaoyu and Gao, Yutong and Yu, Jian and Cao, Yushe and Lei, Xing and Tang, Jinhui},
  journal={arXiv preprint arXiv:2506.01949},
  year={2025}
}

@article{shen2025imaggarment,
  title={IMAGGarment-1: Fine-Grained Garment Generation for Controllable Fashion Design},
  author={Shen, Fei and Yu, Jian and Wang, Cong and Jiang, Xin and Du, Xiaoyu and Tang, Jinhui},
  journal={arXiv preprint arXiv:2504.13176},
  year={2025}
}

@article{shen2024imagpose,
  title={Imagpose: A unified conditional framework for pose-guided person generation},
  author={Shen, Fei and Tang, Jinhui},
  journal={Advances in neural information processing systems},
  volume={37},
  pages={6246--6266},
  year={2024}
}

@inproceedings{gao2024exploring,
  title={Exploring Warping-Guided Features via Adaptive Latent Diffusion Model for Virtual try-on},
  author={Gao, Bo and Ren, Junchi and Shen, Fei and Wei, Mengwan and Huang, Zijun},
  booktitle={2024 IEEE International Conference on Multimedia and Expo (ICME)},
  pages={1--6},
  year={2024},
  organization={IEEE}
}

@inproceedings{shen2024advancing,
title={Advancing Pose-Guided Image Synthesis with Progressive Conditional Diffusion Models},
author={Fei Shen and Hu Ye and Jun Zhang and Cong Wang and Xiao Han and Yang Wei},
booktitle={The Twelfth International Conference on Learning Representations},
year={2024},
url={https://openreview.net/forum?id=rHzapPnCgT}
}

@inproceedings{shen2025boosting,
  title={Boosting consistency in story visualization with rich-contextual conditional diffusion models},
  author={Shen, Fei and Ye, Hu and Liu, Sibo and Zhang, Jun and Wang, Cong and Han, Xiao and Wei, Yang},
  booktitle={Proceedings of the AAAI Conference on Artificial Intelligence},
  volume={39},
  number={7},
  pages={6785--6794},
  year={2025}
}

@ARTICLE{8253599,
  author={Creswell, Antonia and White, Tom and Dumoulin, Vincent and Arulkumaran, Kai and Sengupta, Biswa and Bharath, Anil A.},
  journal={IEEE Signal Processing Magazine}, 
  title={Generative Adversarial Networks: An Overview}, 
  year={2018},
  volume={35},
  number={1},
  pages={53-65},
  doi={10.1109/MSP.2017.2765202}}

@misc{velioglu2025tryoffdiffvirtualtryoffhighfidelitygarment,
      title={TryOffDiff: Virtual-Try-Off via High-Fidelity Garment Reconstruction using Diffusion Models}, 
      author={Riza Velioglu and Petra Bevandic and Robin Chan and Barbara Hammer},
      year={2025},
      eprint={2411.18350},
      archivePrefix={arXiv},
      primaryClass={cs.CV},
      url={https://arxiv.org/abs/2411.18350}, 
}

@article{10.1145/3447239,
author = {Cheng, Wen-Huang and Song, Sijie and Chen, Chieh-Yun and Hidayati, Shintami Chusnul and Liu, Jiaying},
title = {Fashion Meets Computer Vision: A Survey},
year = {2021},
issue_date = {May 2022},
publisher = {Association for Computing Machinery},
address = {New York, NY, USA},
volume = {54},
number = {4},
issn = {0360-0300},
url = {https://doi.org/10.1145/3447239},
doi = {10.1145/3447239},
journal = {ACM Comput. Surv.},
month = jul,
articleno = {72},
numpages = {41}
}

@article{ding2023computational,
  title={Computational technologies for fashion recommendation: A survey},
  author={Ding, Yujuan and Lai, Zhihui and Mok, PY and Chua, Tat-Seng},
  journal={ACM Computing Surveys},
  volume={56},
  number={5},
  pages={1--45},
  year={2023},
  publisher={ACM New York, NY}
}

@inproceedings{shen2025imagdressing,
  title={Imagdressing-v1: Customizable virtual dressing},
  author={Shen, Fei and Jiang, Xin and He, Xin and Ye, Hu and Wang, Cong and Du, Xiaoyu and Li, Zechao and Tang, Jinhui},
  booktitle={Proceedings of the AAAI Conference on Artificial Intelligence},
  volume={39},
  number={7},
  pages={6795--6804},
  year={2025}
}

@inproceedings{choi2021viton,
  title={Viton-hd: High-resolution virtual try-on via misalignment-aware normalization},
  author={Choi, Seunghwan and Park, Sunghyun and Lee, Minsoo and Choo, Jaegul},
  booktitle={Proceedings of the IEEE/CVF conference on computer vision and pattern recognition},
  pages={14131--14140},
  year={2021}
}

@inproceedings{han2018viton,
  title={Viton: An image-based virtual try-on network},
  author={Han, Xintong and Wu, Zuxuan and Wu, Zhe and Yu, Ruichi and Davis, Larry S},
  booktitle={Proceedings of the IEEE conference on computer vision and pattern recognition},
  pages={7543--7552},
  year={2018}
}

@inproceedings{arjovsky2017wasserstein,
  title={Wasserstein generative adversarial networks},
  author={Arjovsky, Martin and Chintala, Soumith and Bottou, L{\'e}on},
  booktitle={International conference on machine learning},
  pages={214--223},
  year={2017},
  organization={PMLR}
}

@inproceedings{zhu2023tryondiffusion,
  title={Tryondiffusion: A tale of two unets},
  author={Zhu, Luyang and Yang, Dawei and Zhu, Tyler and Reda, Fitsum and Chan, William and Saharia, Chitwan and Norouzi, Mohammad and Kemelmacher-Shlizerman, Ira},
  booktitle={Proceedings of the IEEE/CVF conference on computer vision and pattern recognition},
  pages={4606--4615},
  year={2023}
}

@article{ramesh2022hierarchical,
  title={Hierarchical text-conditional image generation with clip latents},
  author={Ramesh, Aditya and Dhariwal, Prafulla and Nichol, Alex and Chu, Casey and Chen, Mark},
  journal={arXiv preprint arXiv:2204.06125},
  volume={1},
  number={2},
  pages={3},
  year={2022}
}

@inproceedings{zhai2023sigmoid,
  title={Sigmoid loss for language image pre-training},
  author={Zhai, Xiaohua and Mustafa, Basil and Kolesnikov, Alexander and Beyer, Lucas},
  booktitle={Proceedings of the IEEE/CVF international conference on computer vision},
  pages={11975--11986},
  year={2023}
}

@article{ye2023ip,
  title={Ip-adapter: Text compatible image prompt adapter for text-to-image diffusion models},
  author={Ye, Hu and Zhang, Jun and Liu, Sibo and Han, Xiao and Yang, Wei},
  journal={arXiv preprint arXiv:2308.06721},
  year={2023}
}

@inproceedings{rombach2022high,
  title={High-resolution image synthesis with latent diffusion models},
  author={Rombach, Robin and Blattmann, Andreas and Lorenz, Dominik and Esser, Patrick and Ommer, Bj{\"o}rn},
  booktitle={Proceedings of the IEEE/CVF conference on computer vision and pattern recognition},
  pages={10684--10695},
  year={2022}
}

@article{dong2024internlm,
  title={Internlm-xcomposer2: Mastering free-form text-image composition and comprehension in vision-language large model},
  author={Dong, Xiaoyi and Zhang, Pan and Zang, Yuhang and Cao, Yuhang and Wang, Bin and Ouyang, Linke and Wei, Xilin and Zhang, Songyang and Duan, Haodong and Cao, Maosong and others},
  journal={arXiv preprint arXiv:2401.16420},
  year={2024}
}

@inproceedings{jo2019sc,
  title={SC-FEGAN: Face editing generative adversarial network with user's sketch and color},
  author={Jo, Youngjoo and Park, Jongyoul},
  booktitle={Proceedings of the IEEE/CVF international conference on computer vision},
  pages={1745--1753},
  year={2019}
}

@inproceedings{lee2020maskgan,
  title={Maskgan: Towards diverse and interactive facial image manipulation},
  author={Lee, Cheng-Han and Liu, Ziwei and Wu, Lingyun and Luo, Ping},
  booktitle={Proceedings of the IEEE/CVF conference on computer vision and pattern recognition},
  pages={5549--5558},
  year={2020}
}

@article{xie2021towards,
  title={Towards scalable unpaired virtual try-on via patch-routed spatially-adaptive gan},
  author={Xie, Zhenyu and Huang, Zaiyu and Zhao, Fuwei and Dong, Haoye and Kampffmeyer, Michael and Liang, Xiaodan},
  journal={Advances in Neural Information Processing Systems},
  volume={34},
  pages={2598--2610},
  year={2021}
}

@inproceedings{dong2019fw,
  title={Fw-gan: Flow-navigated warping gan for video virtual try-on},
  author={Dong, Haoye and Liang, Xiaodan and Shen, Xiaohui and Wu, Bowen and Chen, Bing-Cheng and Yin, Jian},
  booktitle={Proceedings of the IEEE/CVF international conference on computer vision},
  pages={1161--1170},
  year={2019}
}

@inproceedings{gou2023taming,
  title={Taming the power of diffusion models for high-quality virtual try-on with appearance flow},
  author={Gou, Junhong and Sun, Siyu and Zhang, Jianfu and Si, Jianlou and Qian, Chen and Zhang, Liqing},
  booktitle={Proceedings of the 31st ACM International Conference on Multimedia},
  pages={7599--7607},
  year={2023}
}

@inproceedings{morelli2023ladi,
  title={Ladi-vton: Latent diffusion textual-inversion enhanced virtual try-on},
  author={Morelli, Davide and Baldrati, Alberto and Cartella, Giuseppe and Cornia, Marcella and Bertini, Marco and Cucchiara, Rita},
  booktitle={Proceedings of the 31st ACM international conference on multimedia},
  pages={8580--8589},
  year={2023}
}

@article{chong2024catvton,
  title={Catvton: Concatenation is all you need for virtual try-on with diffusion models},
  author={Chong, Zheng and Dong, Xiao and Li, Haoxiang and Zhang, Shiyue and Zhang, Wenqing and Zhang, Xujie and Zhao, Hanqing and Jiang, Dongmei and Liang, Xiaodan},
  journal={arXiv preprint arXiv:2407.15886},
  year={2024}
}

@inproceedings{xu2025ootdiffusion,
  title={Ootdiffusion: Outfitting fusion based latent diffusion for controllable virtual try-on},
  author={Xu, Yuhao and Gu, Tao and Chen, Weifeng and Chen, Arlene},
  booktitle={Proceedings of the AAAI Conference on Artificial Intelligence},
  volume={39},
  number={9},
  pages={8996--9004},
  year={2025}
}

@article{zeng2020tilegan,
  title={TileGAN: category-oriented attention-based high-quality tiled clothes generation from dressed person},
  author={Zeng, Wei and Zhao, Mingbo and Gao, Yuan and Zhang, Zhao},
  journal={Neural Computing and Applications},
  volume={32},
  number={23},
  pages={17587--17600},
  year={2020},
  publisher={Springer}
}

@inproceedings{sun2023sgdiff,
  title={Sgdiff: A style guided diffusion model for fashion synthesis},
  author={Sun, Zhengwentai and Zhou, Yanghong and He, Honghong and Mok, PY},
  booktitle={Proceedings of the 31st ACM international conference on multimedia},
  pages={8433--8442},
  year={2023}
}

@inproceedings{zhang2023diffcloth,
  title={Diffcloth: Diffusion based garment synthesis and manipulation via structural cross-modal semantic alignment},
  author={Zhang, Xujie and Yang, Binbin and Kampffmeyer, Michael C and Zhang, Wenqing and Zhang, Shiyue and Lu, Guansong and Lin, Liang and Xu, Hang and Liang, Xiaodan},
  booktitle={Proceedings of the IEEE/CVF International Conference on Computer Vision},
  pages={23154--23163},
  year={2023}
}

@inproceedings{li2022supervised,
  title={Supervised attribute information removal and reconstruction for image manipulation},
  author={Li, Nannan and Plummer, Bryan A},
  booktitle={European Conference on Computer Vision},
  pages={457--473},
  year={2022},
  organization={Springer}
}

@inproceedings{kwon2022tailor,
  title={Tailor me: An editing network for fashion attribute shape manipulation},
  author={Kwon, Youngjoong and Petrangeli, Stefano and Kim, Dahun and Wang, Haoliang and Swaminathan, Viswanathan and Fuchs, Henry},
  booktitle={Proceedings of the IEEE/CVF Winter Conference on Applications of Computer Vision},
  pages={3831--3840},
  year={2022}
}

@misc{esser2024scalingrectifiedflowtransformers,
      title={Scaling Rectified Flow Transformers for High-Resolution Image Synthesis}, 
      author={Patrick Esser and Sumith Kulal and Andreas Blattmann and Rahim Entezari and Jonas Müller and Harry Saini and Yam Levi and Dominik Lorenz and Axel Sauer and Frederic Boesel and Dustin Podell and Tim Dockhorn and Zion English and Kyle Lacey and Alex Goodwin and Yannik Marek and Robin Rombach},
      year={2024},
      eprint={2403.03206},
      archivePrefix={arXiv},
      primaryClass={cs.CV},
      url={https://arxiv.org/abs/2403.03206}, 
}

@misc{saharia2021imagesuperresolutioniterativerefinement,
      title={Image Super-Resolution via Iterative Refinement}, 
      author={Chitwan Saharia and Jonathan Ho and William Chan and Tim Salimans and David J. Fleet and Mohammad Norouzi},
      year={2021},
      eprint={2104.07636},
      archivePrefix={arXiv},
      primaryClass={eess.IV},
      url={https://arxiv.org/abs/2104.07636}, 
}

@misc{ye2023ipadaptertextcompatibleimage,
      title={IP-Adapter: Text Compatible Image Prompt Adapter for Text-to-Image Diffusion Models}, 
      author={Hu Ye and Jun Zhang and Sibo Liu and Xiao Han and Wei Yang},
      year={2023},
      eprint={2308.06721},
      archivePrefix={arXiv},
      primaryClass={cs.CV},
      url={https://arxiv.org/abs/2308.06721}, 
}

@inproceedings{zhang2018unreasonable,
  title={The unreasonable effectiveness of deep features as a perceptual metric},
  author={Zhang, Richard and Isola, Phillip and Efros, Alexei A and Shechtman, Eli and Wang, Oliver},
  booktitle={Proceedings of the IEEE conference on computer vision and pattern recognition},
  pages={586--595},
  year={2018}
}

@article{ho2022classifier,
  title={Classifier-free diffusion guidance},
  author={Ho, Jonathan and Salimans, Tim},
  journal={arXiv preprint arXiv:2207.12598},
  year={2022}
}

@article{simonyan2014very,
  title={Very deep convolutional networks for large-scale image recognition},
  author={Simonyan, Karen and Zisserman, Andrew},
  journal={arXiv preprint arXiv:1409.1556},
  year={2014}
}

@inproceedings{morelli2022dress,
  title={Dress code: High-resolution multi-category virtual try-on},
  author={Morelli, Davide and Fincato, Matteo and Cornia, Marcella and Landi, Federico and Cesari, Fabio and Cucchiara, Rita},
  booktitle={Proceedings of the IEEE/CVF conference on computer vision and pattern recognition},
  pages={2231--2235},
  year={2022}
}

@article{wang2004image,
  title={Image quality assessment: from error visibility to structural similarity},
  author={Wang, Zhou and Bovik, Alan C and Sheikh, Hamid R and Simoncelli, Eero P},
  journal={IEEE transactions on image processing},
  volume={13},
  number={4},
  pages={600--612},
  year={2004},
  publisher={IEEE}
}

@article{ding2020image,
  title={Image quality assessment: Unifying structure and texture similarity},
  author={Ding, Keyan and Ma, Kede and Wang, Shiqi and Simoncelli, Eero P},
  journal={IEEE transactions on pattern analysis and machine intelligence},
  volume={44},
  number={5},
  pages={2567--2581},
  year={2020},
  publisher={IEEE}
}

@inproceedings{parmar2022aliased,
  title={On aliased resizing and surprising subtleties in gan evaluation},
  author={Parmar, Gaurav and Zhang, Richard and Zhu, Jun-Yan},
  booktitle={Proceedings of the IEEE/CVF conference on computer vision and pattern recognition},
  pages={11410--11420},
  year={2022}
}

@inproceedings{wang2003multiscale,
  title={Multiscale structural similarity for image quality assessment},
  author={Wang, Zhou and Simoncelli, Eero P and Bovik, Alan C},
  booktitle={The thrity-seventh asilomar conference on signals, systems \& computers, 2003},
  volume={2},
  pages={1398--1402},
  year={2003},
  organization={Ieee}
}

@article{binkowski2018demystifying,
  title={Demystifying mmd gans},
  author={Bi{\'n}kowski, Miko{\l}aj and Sutherland, Danica J and Arbel, Michael and Gretton, Arthur},
  journal={arXiv preprint arXiv:1801.01401},
  year={2018}
}

@inproceedings{guo2025any2anytryon,
  title={Any2anytryon: Leveraging adaptive position embeddings for versatile virtual clothing tasks},
  author={Guo, Hailong and Zeng, Bohan and Song, Yiren and Zhang, Wentao and Liu, Jiaming and Zhang, Chuang},
  booktitle={Proceedings of the IEEE/CVF International Conference on Computer Vision},
  pages={19085--19096},
  year={2025}
}

@inproceedings{zhang2023adding,
  title={Adding conditional control to text-to-image diffusion models},
  author={Zhang, Lvmin and Rao, Anyi and Agrawala, Maneesh},
  booktitle={Proceedings of the IEEE/CVF international conference on computer vision},
  pages={3836--3847},
  year={2023}
}

@article{roy2023multi,
  title={Multi-scale attention guided pose transfer},
  author={Roy, Prasun and Bhattacharya, Saumik and Ghosh, Subhankar and Pal, Umapada},
  journal={Pattern Recognition},
  volume={137},
  pages={109315},
  year={2023},
  publisher={Elsevier}
}

@article{xarchakos2024tryoffanyone,
  title={Tryoffanyone: Tiled cloth generation from a dressed person},
  author={Xarchakos, Ioannis and Koukopoulos, Theodoros},
  journal={arXiv preprint arXiv:2412.08573},
  year={2024}
}

@inproceedings{wang2005translation,
  title={Translation insensitive image similarity in complex wavelet domain},
  author={Wang, Zhou and Simoncelli, Eero P},
  booktitle={Proceedings.(ICASSP'05). IEEE International Conference on Acoustics, Speech, and Signal Processing, 2005.},
  volume={2},
  pages={ii--573},
  year={2005},
  organization={IEEE}
}

\end{document}